# Multi-View Learning over Structured and Non-Identical Outputs

**Kuzman Ganchev*** **João V. Graça†** **John Blitzer‡** **Ben Taskar***

*Computer & Information Science
University of Pennsylvania
Philadelphia, PA

†$L^2F$ INESC-ID
INESC-ID
Lisboa, Portugal

‡Natural Language Computing
Microsoft Research Asia
Beijing, China

## Abstract

In many machine learning problems, labeled training data is limited but unlabeled data is ample. Some of these problems have instances that can be factored into multiple views, each of which is nearly sufficent in determining the correct labels. In this paper we present a new algorithm for probabilistic multi-view learning which uses the idea of stochastic agreement between views as regularization. Our algorithm works on structured and unstructured problems and easily generalizes to partial agreement scenarios. For the full agreement case, our algorithm minimizes the Bhattacharyya distance between the models of each view, and performs better than CoBoosting and two-view Perceptron on several flat and structured classification problems.

## 1 Introduction

Multi-view learning refers to a set of semi-supervised methods which exploit redundant views of the same input data (Blum & Mitchell, 1998; Collins & Singer, 1999; Brefeld et al., 2005; Sindhwani et al., 2005). These multiple views can come in the form of context and spelling features in the case of text processing and segmentation, hypertext link text and document contents for document classification, and multiple cameras or microphones in the case of speech and vision. Multi-view methods typically begin by assuming that each view alone can yield a good predictor. Under this assumption, we can regularize the models from each view by constraining the amount by which we permit them to disagree on unlabeled instances. This regularization can lead to better convergence by significantly decreasing the effective size of our hypothesis class (Balcan & Blum, 2005; Kakade & Foster, 2007; Rosenberg & Bartlett, 2007).

In this paper, we propose a probabilistic agreement framework based on minimizing the Bhattacharyya distance (Kailath, 1967) between models trained using different views. Our regularizer is well-suited for optimization of the log-loss, and we give a fast optimization algorithm based on constrained EM (Graca et al., 2008). Where our work is most similar to co-regularization schemes, a minimum Kullbeck-Leibler (KL) distance projection can be computed in closed form resulting in an algorithm that performs better than both CoBoosting and two view Perceptron on several natural language processing tasks. In addition our framework allows us to use different training sets for the two classifiers, even if they have a different label set. In that case, we can reduce the hypothesis space by preferring pairs of models that agree on *compatible* labeling of unlabeled data rather than on *identical* labeling, while still minimizing KL in closed form. When the two views come from models that differ not only in the label set but also in the model structure of the output space, our framework can still encourage agreement, but the KL minimization cannot be computed in closed form. Finally, our method uses soft assignments to latent variables resulting in a more stable optimization procedure.

## 2 Stochastic Agreement

This section outlines stochastic agreement regularization and our method for optimizing it via constrained EM. We can generalize the discussion for multiple views, but for simplicity focus on two views here. We begin by considering the setting of complete agreement. In this setting we have a common desired output for the two models and we believe that each of the two views is sufficiently rich to predict labels accurately. We can leverage this knowledge by restricting our search to model pairs $p_1$,$p_2$ that satisfy

$p_1(y \mid x) \approx p_2(y \mid x)$. Our co-regularized objective is

$$\min_\theta \mathcal{L}_1(\theta_1) + \mathcal{L}_2(\theta_2) + c\mathbf{E}_U[B(p_1(\theta_1), p_2(\theta_2))] \quad (1)$$

where $\mathcal{L}_i = \mathbf{E}[-\log(p_i(y_i|x;\theta_i))] + \frac{1}{\sigma_i^2}||\theta_i||^2$ for $i = 1, 2$ are the standard regularized log likelihood losses of the models $p_1$ and $p_2$, $\mathbf{E}_U[B(p_1, p_2)]$ is the expected Bhattacharyya distance (Kailath, 1967) between the predictions of the two models on the unlabeled data, and $c$ is a constant defining the relative weight of the unlabeled data. The Bhattacharyya distance between two distributions is given by

$$B(p_1, p_2) = -\log \sum_y \sqrt{p_1(y)p_2(y)}.$$

It is a very natural, symmetric measure of distance between distributions which has been used in many signal detection applications (Kailath, 1967). It is also related to the well-known Hellinger distance as well KL-divergence as we outline below.

### 2.1 Optimization algorithm

In order to optimize the objective we first consider a variational definition of Bhattacharyya distance that will allow us to generalize our approach to structured and non-identical output spaces. Proposition 2.1 relates the Bhattacharyya regularization term to the value of a constrained minimum KL problem, where the constraints enforce agreements between the two views. Proposition 2.2 shows that a simple majorization-minimziation algorithm optimizes the desired objective. The proof is in the appendix.

**Proposition 2.1** *The Bhattacharyya distance* $-\log \sum_y \sqrt{p_1(y)p_2(y)}$ *is equal to* $\frac{1}{2}$ *of the value of the convex optimization problem*

$$\min_{q \in \mathcal{Q}} \quad \mathrm{KL}(q(y_1, y_2)||p_1(y_1)p_2(y_2))$$
$$\text{where} \quad \mathcal{Q} = \{q : \mathbf{E}_q[\delta(y_1 = y) - \delta(y_2 = y)] = 0 \ \forall y\},$$

*where* $\delta(cond)$ *is* 1 *if cond is true and* 0 *otherwise. Furthermore, the minimizer decomposes as* $q(y_1, y_2) = q_1(y_1)q_2(y_2)$ *and is given by* $q_i(y) \propto \sqrt{p_1(y)p_2(y)}$.

Replacing the Bhattacharyya regularization term in Equation 1 with the program of Proposition 2.1 yields the objective

$$\min_\theta \mathcal{L}_1(\theta) + \mathcal{L}_2(\theta) + c\mathbf{E}_U\left[\min_{q\in\mathcal{Q}(\mathbf{x})} \mathrm{KL}(q(y_1 y_2) \,||\, p(y_1)p(y_2))\right].$$

Note that this objective is convex separately in $\theta$ and $q$, but not jointly. We can optimize it iteratively using the constrained EM framework of Graca et al. (2008). We define $\mathbf{agree}(p_1, p_2)$ to be the minimizer of Proposition 2.1, and present the optimization in Algorithm 1.

**Algorithm 1** minimizes co-regularized loss:

$$\mathcal{L}_1(\theta) + \mathcal{L}_2(\theta) + c\mathbf{E}_U[\min_{q\in\mathcal{Q}(\mathbf{x})} \mathrm{KL}(q(y_1, y_2) \,||\, p_1(y_1)p_2(y_2)].$$

1: $\theta_1 \leftarrow \min_\theta \mathcal{L}_1(\theta_1)$
2: $\theta_2 \leftarrow \min_\theta \mathcal{L}_2(\theta_2)$
3: **for** $n$ iterations **do**
4: $\quad q(y_1, y_2|\mathbf{x}) \leftarrow \mathbf{agree}(p_1(y_1|x), p_2(y_2|x)) \ \ \forall x \in U$
5: $\quad \theta_1 \leftarrow \min_\theta \mathcal{L}_1(\theta) - c \mathop{\mathbf{E}}_{x\sim U, y_1\sim q}[\log p_1(y_1|x;\theta)]$
6: $\quad \theta_2 \leftarrow \min_\theta \mathcal{L}_2(\theta) - c \mathop{\mathbf{E}}_{x\sim U, y_2\sim q}[\log p_2(y_2|x;\theta)]$
7: **end for**

**Proposition 2.2** *Fixpoints of the loop in Algorithm 1 are local minima of*

$$\mathcal{L}_1(\theta_1) + \mathcal{L}_2(\theta_2) + c\mathbf{E}_U\left[\min_{q\in\mathcal{Q}(\mathbf{x})} \mathrm{KL}(q(y_1, y_2) \,||\, p(y_1)p(y_2))\right].$$

If we consider the labels of the unlabeled data to be hidden variables, then Algorithm 1 is the constrained EM algorithm of Graca et al. (2008). Line 4 corresponds to the constrained E-step, while lines 5 and 6 constitute the M-step of the algorithm. The proof of Proposition 2.2 is a straightforward modification of the constrained EM proof in Graca et al. (2008).

Recall that we defined $\mathbf{agree}(p_1, p_2)$ to be the minimizer of the convex program in Proposition 2.1. The same proposition gives us a simple closed-form solution for this minimizer. In Section 2.2 shows that this solution is also easy to compute for structured models. We then show in Section 2.3 how a simple change to the constraints of the convex program allows us to optimize a natural co-regularizer in the partial agreement setting.

### 2.2 Undirected graphical models

Our regularizer extends to full agreement for undirected graphical models. In the case where $p_1$ and $p_2$ have the same structure, $q = \mathbf{agree}(p_1, p_2)$ will share this structure and the projection can be computed in closed form. Let $p_1(y \mid x) = Z_1^{-1} \prod_c \phi_1(y_c, x)$, where $c$ is a clique in the graphical model and $\phi_1(y_c, x)$ is the corresponding clique potential, similarly for $p_2$. The derivation follows from the fact that $q(y_1, y_2)$ factors into a product of $q_1(y_1) = q_2(y_2)$, such that:

$$q_i(y)^2 \propto p_1(y)p_2(y) \quad (2)$$
$$= Z_1^{-1} Z_2^{-1} \prod_c \phi_1(y_c)\phi_2(y_c) \quad (3)$$
$$= Z_q^{-2} \prod_c \phi_q(y_c)^2, \quad (4)$$

where $\phi_q(y_c) = \sqrt{\phi_1(y_c)\phi_2(y_c)}$. This means that we can represent $q$ using the same cliques as $p_1$ and $p_2$, and the clique potentials for $q$ are just the square root of the product of the clique potentials for $p_1$ and $p_2$. In the case of log-linear Markov random fields, the clique potentials are stored in log space so this corresponds to averaging the values before computing normalization.

### 2.3 Partial agreement and hierarchical labels

Our method extends naturally to partial-agreement scenarios. For example we can encourage two part of speech taggers with different tag sets to produce compatible parts of speech, such as noun in tag set one and singular-noun in tag set 2 rather than noun in tag set 1 and verb in tag set 2. In particular, suppose we have a mapping from both label sets into a common space where it makes sense to encourage agreement. For the part of speech tagging example, this could mean mapping all nouns from both tagsets into a single class, all verbs into another class and so on. In general, we might have functions $g_1(y_1)$ and $g_2(y_2)$ that map variables for the two models onto the same space. We want to encourage agreement with $p_1(z) \approx p_2(z)$ where $p_1(z) = \sum_y \delta(g_1(y) = z)p_1(y)$ and similarly for $p_2$. Also, denote $p_i(y_i|z) = p_i(y_i)/p_i(z)$. In this case, our objective becomes

$$\min_{\theta} \mathcal{L}_1(\theta_1) + \mathcal{L}_2(\theta_2) + c\mathbf{E}_U[B(p_1(z), p_2(z))] \ . \quad (5)$$

In the special case where some labels are identical for the two models and others are incompatible, we have $g_1(y_1)$ mapping the incompatible labels into one bin and the others into their own special bins.

By slightly changing the constraints to the program in Proposition 2.1. With slight abuse of notation we denote $g_1(y_1) = z$ as $y_1 \mapsto z$. The following proposition allows us to optimize equation 5.

**Proposition 2.3** *The Bhattacharyya distance* $-\log \sum_z \sqrt{p_1(z)p_2(z)}$ *is* $\frac{1}{2}$ *the value of the convex program*

$$\begin{array}{cc} \min\limits_{q \in \mathcal{Q}} & \mathrm{KL}(q(y_1, y_2)||p_1(y_1)p_2(y_2)) \\ \text{where} & \mathcal{Q} = \{q : \mathbf{E}_q[\delta(y_1 \mapsto z) - \delta(y_2 \mapsto z)] = 0 \ \forall z\} \ . \end{array}$$

*Furthermore, the minimizer decomposes as* $q(y_1, y_2) = q_1(y_1|z_1)q_1(z_1)q_2(y_2|z_2)q_2(z_2)$, *where* $q_1(z_1) = q_2(z_2) \propto \sqrt{p_1(z_1)p_2(z_2)}$ *and* $q_1(y_1|z_1) = p_1(y_1|z_1)$ *and* $q_2(y_2|z_2) = p_2(y_2|z_2)$.

By using the program in Proposition 2.3 instead of the one in Proposition 2.1 as the distribution $\mathbf{agree}(p_1, p_2)$ in Algorithm 1) we optimize the partial agreement objective using Algorithm 1. Note in particular that the computation of $\mathbf{agree}(p_1, p_2)$ is still in closed form.

### 2.4 Partial agreement in structured problems

For the unstructured case, it is easy to compute $p(z)$. Unfortunately if we collapse some labels, $p(z)$ might not have the same Markov properties as $p(y)$. For example, if $p$ is a distribution over three states (1,2,3) that assigns probability 1 to the sequence (1,2,3,1,2...) and probability zero to other sequences. This is a first-order Markov chain. If the mapping is $1 \mapsto 1$ and $2, 3 \mapsto 0$ then $p(z)$ assigns probability 1 to (1,0,0,1,0...), which cannot be represented as a first-order Markov chain. Essentially, the original chain relied on being able to distinguish between the allowable transition (2,3) and the disallowed transition (3,2). When we collapse the states, both of these transitions map to (0,0) and cannot be distinguished. Consequently, the closed form solution given in Proposition 2.3 is not usable. Potentially, we could compute some approximation to $p(z)$ and from that compute an approximation to $q$. Instead, we re-formulate our constraints for the structured case and use conjugate gradient to find the optimal $q$. We ensure that the proposal distribution will have the same form as $p_1$ and $p_2$ by requiring only that the marginals of each clique match the desired marginals rather than requiring the joint to have the same probability:

$$\begin{array}{cc} \min\limits_{q} & \mathrm{KL}(q(y_1, y_2)||p_1(y_1)p_2(y_2)) \\ \text{s.t.} & \mathbf{E}_q[\delta(y_{1,c} \mapsto z_c) - \delta(y_{2,c} \mapsto z_c)] = 0 \ \forall z_c. \end{array} \quad (6)$$

In our experiments we found that 10 gradient steps are sufficient for practical purposes.

## 3 Relation to previous work

There has been a significant amount of work on two view learning in the past few years. Co-Training (Blum & Mitchell, 1998) labels increasingly large amounts of unlabeled data with a classifier from each view and re-trains each classifier on the unlabeled data. If the two views are generated independently of each other given the label, Co-Training is guaranteed to work. Nigam & Ghani (2000) find that this assumption is violated in practice and that Co-Training performance suffers significantly. Pierce & Cardie (2001) empirically investigate Co-Training for noun phrase chunking and find the same problem. They suggest correcting by hand the examples labeled by the classifiers. Unlike Co-Training, CoBoosting (Collins & Singer, 1999) and two view Perceptron (Brefeld et al., 2005) re-label the unlabeled data at each iteration. Potentially this might lead to a more stable optimization algorithm, since the identity of the first few labeled examples has a smaller effect on the outcome. Of course directly optimizing the objective would lead to even more stability. Sindhwani et al. (2005) introduce co-regularized least squares and co-regularized SVMs and directly optimize the objective functions they propose. Like them, we know what our objective function is and our optimization procedure is guaranteed to find a local optimum. Kakade & Foster (2007) investigate two view linear regression and show that the hypothesis space can be reduced by performing CCA on input data.

In the case where we compute the KL projection in closed form, we are essentially using the two classifiers as a logarithmic opinion pool (Bordley, 1982). Smith et al. (2005) find that using logarithmic opinion pools of unregularized conditional random fields performs as well as regularized CRFs, and does not require tuning a regularization parameter. This suggests that co-regularized CRFs should be less sensitive to regularization hyper-parameters than a monolithic CRF. Suzuki et al. (2007) use a logarithmic opinion pool of several discriminatively trained CRFs and several hidden Markov models for named entity recognition and syntactic chunking. Computing optimal mixing weights on a held-out portion of the training set, they iteratively re-train the HMMs on the data labeled with the opinion pool. Since their optimization algorithm is very similar to ours, we can derive an objective function for their optimization from our work.

There is an entire class of different semi-supervised methods which assume that the true decision boundary lies on low density regions of the input space Chapelle et al. (2006). For example, Jiao et al. (2006) minimizes entropy on unlabeled data for CRFs. Mann & McCallum (2007) is also quite related, as it regularizes feature expectations on unlabeled data.

In the next section we empirically compare our method with CoBoosting and two view Perceptron. Since these methods are based on different objective functions it is worth examining where each one works best. Altun et al. (2003) compare log-loss and exp-loss for sequential problems. They find that the loss function does not have as great an effect on performance as the feature choice. However, they also note that exp-loss is expected to perform better for clean data, while log-loss is expected to perform better when there is label noise. The intuition behind this is in the rate of growth of the loss functions. Exp-loss grows exponentially with misclassification margin while log-loss grows linearly. Consequently when there is label noise, AdaBoost focuses more on modeling the noise. Since Co-Boosting optimizes a co-regularized exp-loss while our work optimizes a co-regularized log-loss we expect to do better on problems where the labels are noisy.

To get more intuition about this, Figure 1 shows the co-regularization loss functions for our method (top), CoBoosting (middle) and co-RLS (bottom). For two underlying binary linear classifiers, $\hat{y} = \text{sign}(w \cdot x)$, the horizontal axes represent the value of the dot product, while the vertical axis is the loss. If we consider the plane parallel to the page, we see how the different co-regularizers penalize the classifiers when they disagree and are equally confident in their decision. Restricted to this plane, all three co-regularizers grow at the same asymptotic rate as the loss functions for the individual models: Linearly for our work, exponentially for Co-Boosting and quadratically for co-RLS. If we look at the area where the two models agree (the flat part of the Co-Boosting graph) we see what the penalty is when the classifiers agree but have different confidence. In this case co-RLS is harshest since it penalizes differences in the dot product equally regardless of the absolute value of the dot product. Intuitively, this is a problem. If one model predicts 1 with confidence 0.5 and the other predicts -1 with confidence 0.5 they are disagreeing while if they both predict 1 with confidence 1000 and 1001 respectively, they are agreeing on the label and are very close in their confidence estimates. At the other extreme, Co-Boosting imposes almost no penalty whenever the two classifiers agree, regardless of their confidence. The Bhattacharyya distance co-regularizer lies between these extremes, penalizing differences in confidence near the origin but is more lenient when the classifiers are both very confi-dent and agree.

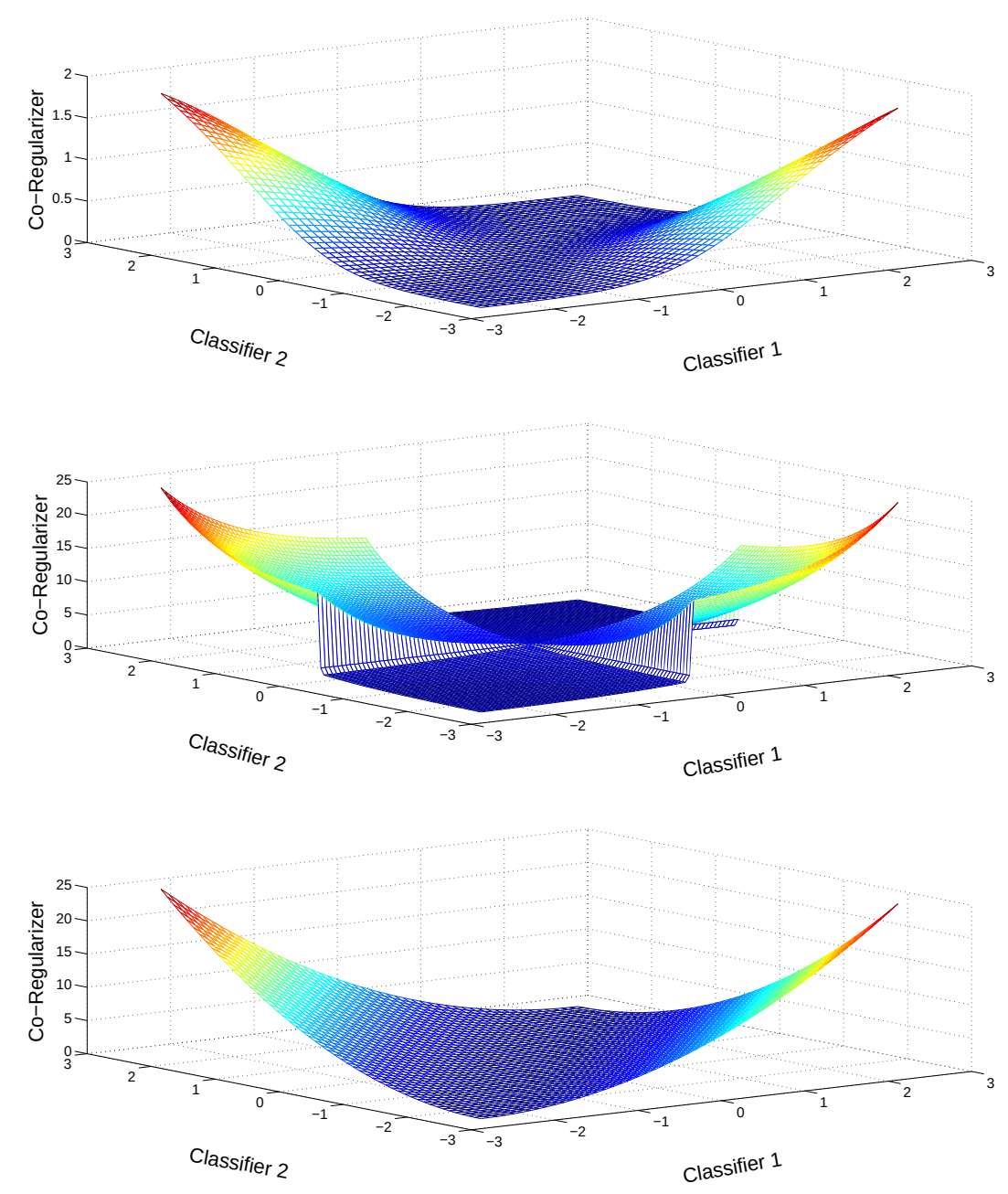

Figure 1: Different Loss Functions. Top: Bhattacharyya distance regularization. Middle: Exp-loss regularization. Bottom: Least Square regularization

We compare our work with two view Perceptron for both structured and unstructured problems. Altun et al. (2002) describe an extension of AdaBoost to structured problems and an extension of CoBoosting is immediate, but we did not implement it due to time constraints.

Finally, if we have labeled data from one domain but want to apply it to another domain we can use any of the co-training frameworks mentioned earlier, including our own to perform domain transfer. For sentiment classification we will see that our method performs comparably with Structural Correspondence Learning (Blitzer et al., 2006), which is based on Alternating Structure Optimization (Ando & Zhang, 2005).

## 4 Experiments

### 4.1 Classification problems

Our first set of experiments is for transfer learning for sentiment classification. We used the data from Blitzer et al. (2007). The two views are generated from a random split of the features. We compare our method to several supervised methods as well as Co-Boosting (Collins & Singer, 1999), two view Perceptron (Brefeld et al., 2005) and structural correspondence learning (Blitzer et al., 2007). The column labeled "SCL" contains the best results from Blitzer et al. (2007), and is not directly comparable with the other methods since it uses some extra knowledge about the transfer task to choose auxiliary problems. For all the two-view methods we weigh the total labeled data equally with the total unlabeled data. We regularize the maximum entropy classifiers with a unit variance Gaussian prior. The results are in Table 1. Our method is labeled "SAR" (Stochastic Agreement Regularization). Out of the 12 transfer learning task, our method performs best in 6 cases, SCL in 4, while CoBoosting performs best only once. Two view Perceptron never outperforms all other methods. One important reason for the success of our method is the relative strength of the maximum entropy classifier relative to the other supervised methods for this particular task. We expect that Co-Boosting will perform better than our method in situations where Boosting significantly outperforms maximum entropy.

The next set of experiments we performed was with named entity disambiguation. Given a set of split named entities, we want to predict what type of named entity each one is. We use the training data from the 2003 CoNLL shared task (Sang & Meulder, 2003). The two views comprise content versus context features. The content features were words, POS tags and character n-grams of length 3 for all tokens in the named entity, while context features the same but for three words before and after the named entity. We used 2000 examples as testing data and roughly 30,000 as unlabeled data. Table 2 shows the results for different amounts of training data. For this dataset, we choose the variance of the Gaussian prior as well as the relative weighting of the labeled and unlabeled data by cross validation on the training set. In order to test whether the advantage our method gets is from the joint objective or from the use of $\mathbf{agree}(p_1, p_2)$, which is an instance of logarithmic opinion pools, we also report the performance of using $\mathbf{agree}(p_1, p_2)$ when the two views $p_1$ and $p_2$ have been trained only on the labeled data. In the column labeled "$\mathbf{agree}_0$" we see that for this dataset the benefit of our method comes from the joint objective function rather than from the use of logarithmic opinion pools.

| Data size | mx-ent | $\mathbf{agree}_0$ | SAR | RRE |
|---|---|---|---|---|
| 500 | 74.0 | 74.4 | 76.4 | 9.2% |
| 1000 | 80.0 | 80.0 | 81.7 | 8.5% |
| 2000 | 83.4 | 83.4 | 84.8 | 8.4% |

Table 2: Named entity disambiguation. Prior variance and $c$ chosen by cross validation. $\mathbf{agree}_0$ refers to performance of two view model before first iteration of EM. RRE is reduction in error relative to error of MaxEnt model.

### 4.2 Partial Agreement

We also wanted to investigate the applicability of this method to the partial agreement setting. Suppose that you are interested in a classification task and you have labeled a small training corpus. You also have available a training corpus from some other source that has some of the distinctions you are interested in, but not others. You could use our method by training a classifier for your task as well as one for the auxiliary data and encouraging them to agree on the appropriate labels on unlabeled data. To investigate this scenario, we chose four newsgroups from the 20-newsgroups corpus (Lang, 1995), namely comp.sys.ibm.pc.hardware, comp.sys.mac.hardware, talk.politics.mideast, rec.sport.baseball. Our views consist of one model trained on a subset of the newsgroups data with all four labels and another model trained on a subset where talk.politics.mideast and rec.sport.baseball have been collapsed into one category. Figure 2 shows the average accuracies over 10 runs with differing amounts of training data. In all cases, the number of labeled examples per view are equal and we regularize using a Gaussian prior with variance 10. The labeled and unlabeled data have equal weighting for these experiments. We see in the

| Domains | MIRA | Boost | Perc | mx-ent | SCL | CoBoost | coPerc | SAR |
|---|---|---|---|---|---|---|---|---|
| books→dvds | 77.2 | 72.0 | 74 | 78.5 | 75.8 | 78.8 | 75.5 | **79.8** |
| dvds→books | 72.8 | 74.8 | 74.5 | 80.3 | 79.7 | 79.8 | 74.5 | **81.3** |
| books→electr | 70.8 | 70.3 | 73.3 | 72.5 | 75.9 | **77.0** | 69.3 | 75.5 |
| electr→books | 70.7 | 62.5 | 73 | 72.8 | **75.4** | 71.0 | 67.5 | 74.3 |
| books→kitchn | 74.5 | 76.3 | 73.5 | 77.8 | 78.9 | 78.0 | 76.5 | **81.0** |
| kitchn→books | 70.9 | 66.5 | 67.3 | 70.3 | 68.6 | 69.8 | 66 | **72.8** |
| dvds→electr | 73.0 | 73.2 | 73.5 | 75.5 | 74.1 | 75.3 | 71.2 | **76.5** |
| electr→dvds | 70.6 | 66.3 | 64.8 | 69.3 | **76.2** | 73.5 | 63.3 | 73.0 |
| dvds→kitchn | 74.0 | 75.5 | 78.3 | 80.5 | 81.4 | 79.0 | 78.25 | **82.8** |
| kitchn→dvds | 72.7 | 61.8 | 64 | 69.5 | **76.9** | 70.1 | 60.5 | 72.8 |
| electr→kitchn | 84.0 | 73.2 | 81 | **86.5** | 85.9 | 85.0 | 83.3 | 85.8 |
| kitchn→electr | 82.7 | 66.3 | 81 | 82.8 | **86.8** | 83.0 | 80.5 | 85.5 |

Table 1: Performance of several methods on a sentiment classification transfer learning task. Reviews of objects of one type are used to train a classifier for reviews of objects of another type. The abbreviations in the column names are as follows. Boost: AdaBoost algorithm, Perc: Perceptron, mx-ent: maximum entropy, SCL: structural correspondence learning, CoBoost: CoBoosting, coPerc: two view Perceptron, SAR: this work. The best accuracy is shown in bold for each task.

figure that while SAR performs better than the supervised model, most of the benefit comes from agreement during inference (Agree 0 curve in Figure 2).

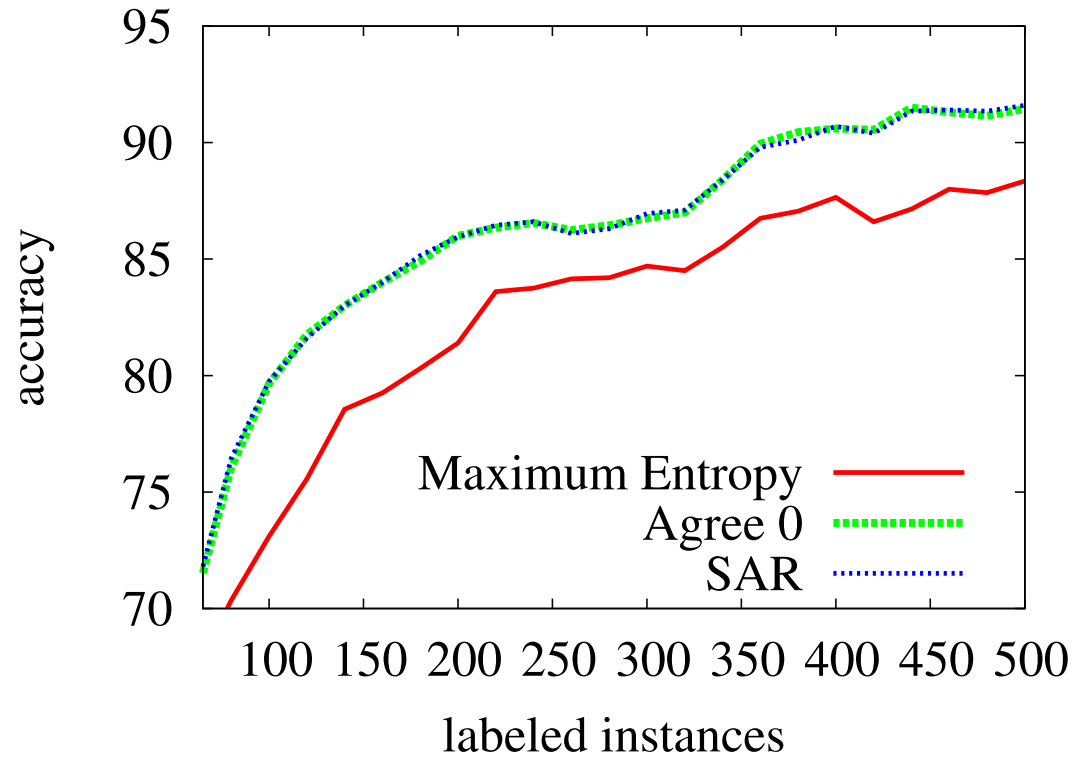

Figure 2: Results of the partial agreement experiments with four labels in one view and three labels in the other.

Table 3 shows the average confusion matrix for the runs with 500 training examples. Rows represents correct labels, while the columns represent labels assigned by the model. Each cell contains the percentage of all instances with the given correct label for the fully supervised case on top, and the difference to the same value for the partial agreement case below. We see that the auxiliary partially labeled data helps us to reduce the confusion between the difficult comp.sys.ibm.pc.hardware, comp.sys.mac.hardware categories. We see a similar trend in confusion matrices for all amounts of training data.

| | pc | mac | pol | bball |
|---|---|---|---|---|
| pc | 82.9 | 16.5 | 0 | 0.5 |
| | **5.1** | **−5.6** | 0.4 | 0.2 |
| mac | 15.0 | 83.7 | 0.2 | 1.2 |
| | **−3.7** | **3.6** | **−0.2** | 0.1 |
| politics | 1.4 | 2.3 | 93.2 | 3.2 |
| | **−0.7** | **−1.2** | **1.6** | 0.2 |
| bball | 1.9 | 2.5 | 0.6 | 95.1 |
| | **−0.7** | **−1.9** | 0.2 | **2.2** |

Table 3: Part of the confusion matrix for the partial agreement scenario. Top in each cell: Percents for fully supervised. Bottom in each cell: Percent difference to partial agreement. Bold represents improvement. See text for description.

### 4.3 Structured models

In order to investigate the applicability of our method to structured learning we apply it to the shallow parsing task of noun phrase chunking. We performed our experiments on the English training portion of the CoNLL 2000 shared task (Sang & Buchholz, 2000). We selected 500 sentences as testing data, varying amounts of data for training and the remainder was used as unlabeled data. We use content and context views, where the content view is the current word and POS tag while the context view is the previous and next words and POS tags. We regularize the CRFs with a variance 10 Gaussian prior and weigh the unlabeled data so that it has the same total weight as the labeled data. The variance value was chosen based on preliminary experiments with the data. Table 4 shows the F-1 scores of the different models. We compare our method to a monolithic CRF as well as av-eraged Perceptron the two view Perceptron of Brefeld et al. (2005) with averaging. The Perceptron models were trained for 20 iterations. Preliminary experiments show that performance on held out data does not change after 10 iterations so we believe the models have converged. Both two view semi-supervised methods show gains over the corresponding fully-supervised method for 10-100 sentences of training data, but do not improve further as the amount of labeled data increases. The method presented in this paper outperforms two view Perceptron when the amount of labeled data is very small, probably because regularized CRFs perform better than Perceptron for small amounts of data. As the number of training sentences increases, two view Perceptron performs as well as our method, but at this point it has little or no improvement over the fully-supervised Perceptron.

| size | CRF | SAR(RRE) | Perc | coPerc |
|---|---|---|---|---|
| 10 | 73.2 | **78.2** (19%) | 69.4 | 71.2 |
| 20 | 79.4 | **84.2** (23%) | 74.4 | 76.8 |
| 50 | 86.3 | **86.9** (4%) | 80.1 | 84.1 |
| 100 | 88.5 | **88.9** (3%) | 86.1 | 88.1 |
| 200 | 89.6 | 89.6 (0%) | 89.3 | **89.7** |
| 500 | **91.3** | 90.6 (-8%) | 90.8 | 90.9 |
| 1000 | 91.6 | 91.1 (-6%) | 91.5 | **91.8** |

Table 4: F-1 scores for noun phrase chunking with context/content views. Testing data comprises 500 sentences, with 8436 sentences divided among training and unlabeled data. The best score is shown in bold for each training data size.

### 4.4 Structured partial agreement

Finally, we ran some named entity recognition experiments where the two views had different label sets. All our experiments use the training portion of the 2003 CoNLL shared task (Sang & Meulder, 2003), regularize the model parameters with a variance 10 Gaussian prior and weigh the labeled and unlabeled data equally. The view of primary interest had segmentation of names in the categories "person", "location", "organization" and "miscellaneous". The auxiliary view collapsed the "location" and "miscellaneous" labels. For the experiments we use a total of about 14 thousand sentences, of which 200 are used for testing, and the rest are split between training for each of the two models and "unlabeled" data. The F-1 score as we vary the amount of training data available to each model are shown in Figure 3. The number of training sentences are the same for both views and are shown on the horizontal axis. We see that the semi-supervised model does better than the baseline in most cases, but most of the improvement is from the combination of the two classifiers rather than from joint learning.

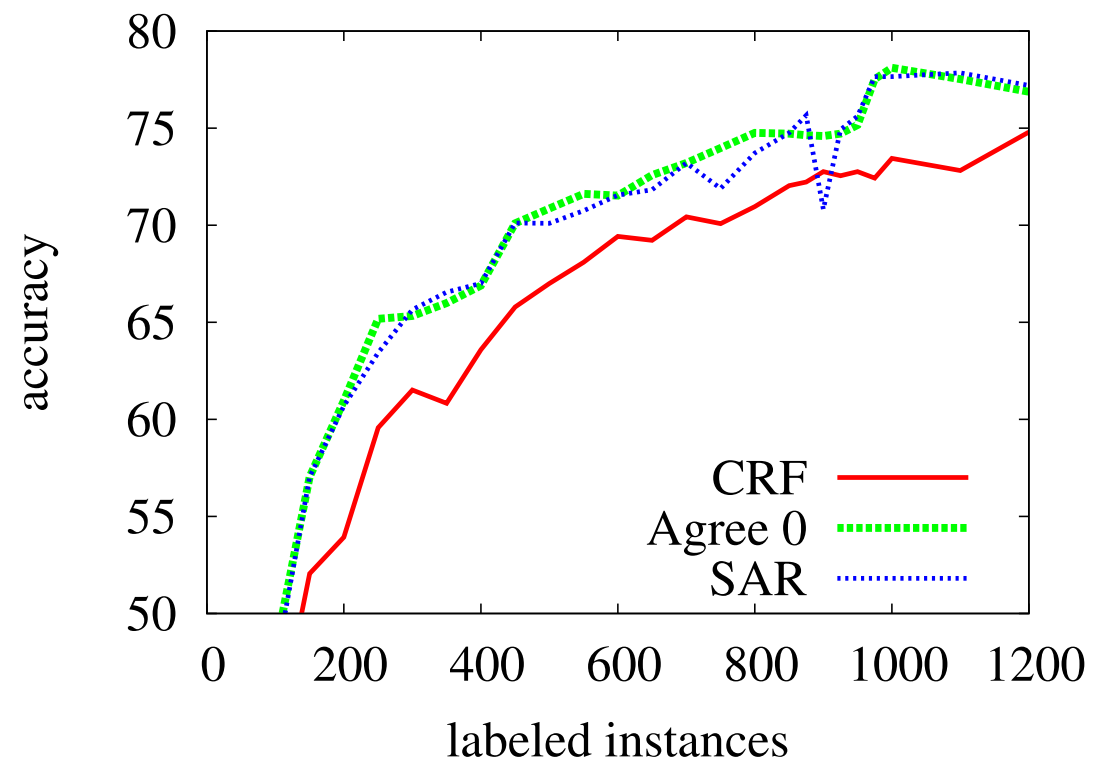

Figure 3: Results of the partial agreement experiments for structure output on name entity resolution, with all (four) categories in one view and three categories in the other (Location and Misc where collapsed into misc).

## 5 Conclustion and Future Work

We have introduced a novel two-view co-regularization appropriate for probabilistic models, which naturally extends to multiple views. In the normal two-view setting where the output spaces of the two views are identical, the co-regularization penalty is the Bhattacharyya distance. Our framework extends naturally to structured problems and also to partial agreement scenarios. We compared the framework with CoBoosting and two view Perceptron as well as a state of the art transfer learning algorithm, and found that our method often outperforms all other methods. Additionally, we demonstrate that our framework can be used in cases where the output spaces of the two views are not identical and the other methods are not directly applicable. A natural extension of this work would be to encourage agreement between more than two views, and to weigh the views differently as in logarithmic opinion pools. Another direction for future work is the application of this framework to problems where the proposal distribution cannot be computed in closed form, such as combining a dependency parser with a phrase structure parser. Finally, it would be interesting to investigate under what conditions two view co-regularization frameworks such as ours can be expected to work. For example, under what conditions should we expect the two models to converge asymptotically faster than a monolithic model?

### Acknowledgements

Kuzman Ganchev was partially supported by NSF ITR EIA 0205448. João V. Graça was supported by a fellowship from Fundação para a Ciência e Tecnologia (SFRH/ BD/ 27528/ 2006).

## A Proof of proposition 2.1

Taking the dual of the optimization problem in Equation 2.1 we get

$$\arg\max_{\lambda} - \log \sum_{y_1, y_2} p(y_1, y_2) \exp(\lambda \cdot \psi) \qquad (7)$$

with $q(y_1, y_2) \propto p(y_1, y_2) \exp(\lambda \cdot \psi(y_1, y_2))$. Where $\psi(y_1, y_2)$ is a vector of features of the form $\delta(y_1 = y) - \delta(y_2 = y)$ with one entry for each possible label $y$. Noting that the features decompose into $\psi'(y_1) - \psi'(y_2)$, we know that $q(y_1, y_2)$ decomposes as $q_1(y_1)q_2(y_2)$. Furthermore, our constraints require that $q_1(y) = q_2(y) \forall y$ so we have $q(y_1)q(y_2) \propto p_1(y_1)\exp(\lambda \cdot \psi'(y_1))p_2(y_2)\exp(-\lambda \cdot \psi'(y_2))$. Letting $y_1 = y_2$ we have $q(y)^2 = p_1(y)p_2(y)$ which gives us a closed form computation of $\mathbf{agree}(p_1, p_2) \propto \sqrt{p_1(y)p_2(y)}$. Substituting this solution into the program of Proposition 2.1, and performing algebraic simplification yields the desired result.